\pdfoutput=1
\documentclass{article}

\PassOptionsToPackage{numbers,sort&compress}{natbib}
\usepackage[preprint]{neurips_2026}

\usepackage[utf8]{inputenc}
\usepackage[T1]{fontenc}
\usepackage{graphicx}
\usepackage{amsmath}
\usepackage{amssymb}
\usepackage{booktabs}
\usepackage{microtype}
\usepackage{xcolor}
\usepackage{url}
\usepackage{flafter}
\usepackage{placeins}
\usepackage{capt-of}
\usepackage[hidelinks]{hyperref}
\hypersetup{
  pdftitle={Grounding in the Dark: Text-Only Training for Zero-Shot Video Temporal Localization},
  pdfauthor={Divyanshu Mishra, Sabine Sternig, Rakshith Shetty, Erhan Gundogdu}
}

\title{Grounding in the Dark: Text-Only Training for Zero-Shot Video Temporal Localization}

\author{%
  Divyanshu Mishra\textsuperscript{1,2}\thanks{Most of this work was done during an internship at Amazon Berlin (August--December 2024).} \quad
  Sabine Sternig\textsuperscript{1} \quad
  Rakshith Shetty\textsuperscript{1} \quad
  Erhan Gundogdu\textsuperscript{1} \\
  \textsuperscript{1}Amazon \quad \textsuperscript{2}University of Oxford
}

\begin{document}
\raggedbottom
\maketitle

\begin{abstract}
Video-LLMs are typically trained on paired video-text data for temporal tasks,
yet collecting precise temporal annotations and training on large-scale video
datasets are both costly. We investigate whether a text-only LLM, with no
vision-language pretraining, can be trained on textual video descriptions alone
to perform temporal grounding and receive video only at inference time. This
setting removes paired video data from training altogether and is a step toward
training multimodal models without paired annotations. We generate a 134K-example
corpus of textual sequences with up to 100 time steps, adapt Qwen3-8B with LoRA
on CLIP or LLM2CLIP text embeddings, and evaluate on Charades-STA and
QV-Highlights without training on videos from either benchmark. LLM2CLIP, whose
text encoder supports longer inputs, represents detailed descriptions better
than CLIP (61.54 versus 53.86 mtIoU on Temporal-TextVid) and transfers them
better to video. Still, replacing text with frame embeddings at inference yields
only 4.99 mtIoU on Charades-STA and 11.12 on QV-Highlights, revealing a large
modality gap. To bridge it, we propose Nearest Entity Replacement (NER), a
training-free mapping that represents each video frame by the normalized mean of
its nearest text-entity embeddings, so that the model receives the kind of input
it was trained on. NER reaches 14.45 mtIoU on Charades-STA and 20.72 on
QV-Highlights, outperforming the strongest of four families of modality-gap
baselines by 5.00 and 1.71 points. NER thus enables partial transfer from text
to video, while more reliable grounding will likely require encoders with
better-aligned text and visual representations.
\end{abstract}

\section{Introduction}
\begin{figure}[t]
    \centering
    \includegraphics[trim=0 360 0 0,clip,width=0.92\linewidth]{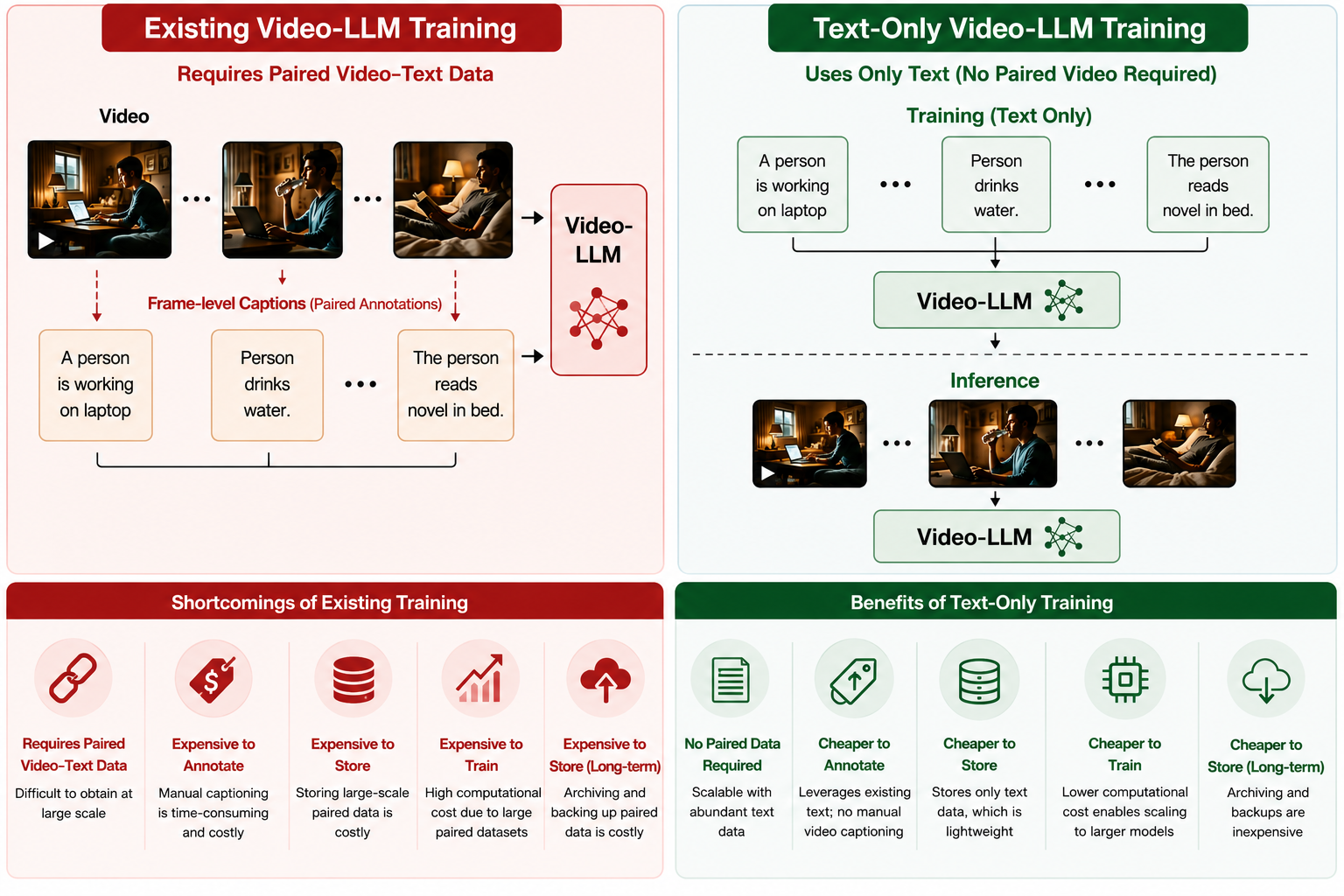}
    \caption{Paired video-text training versus our text-only training and video-inference setting.}
    \label{fig:problem_intro}
\end{figure}

Video large language models (Video-LLMs) can describe videos, answer questions
about them, and localize events in time
\cite{li2023videochat,ren2024timechat,huang2024vtimellm,wang2024hawkeye}.
Recent trained temporal-grounding variants encode time with dedicated tokens
\cite{wang2025grounded}, numbered frames \cite{wu2025numberit}, sequential
transfer from images and short clips \cite{deng2025seq2time}, or
reinforcement-learning objectives \cite{chen2025datasets}. Although their
architectures differ, these systems learn from paired video and language.
Marking event boundaries requires close inspection by annotators, while storing
and repeatedly decoding long videos adds training cost
(Fig.~\ref{fig:problem_intro}).

We ask whether a language model can learn a video task from text alone and
still operate on video. We start from the Qwen3-8B language checkpoint rather
than a pretrained Video-LLM and adapt it with LoRA
\cite{qwen3_2025,hu2022lora}. A frozen contrastive encoder
provides the only pre-existing image-text alignment. During training, its text
encoder processes one description per time step; at inference, the
corresponding visual encoder processes sampled frames. The downstream
projection and LLM remain unchanged. This setup tests whether a temporal model
trained on text embeddings can use visual embeddings without retraining.

 Temporal localization provides a demanding transfer test because the query
specifies \emph{what} event to find but not \emph{when}. Some VideoQA
datasets admit language shortcuts or strong text-only performance
\cite{yang2021just,sadhu2021phrases,korkut2026visual}, whereas our setting
replaces every textual position with a frame embedding at inference. The model must find both boundaries among up to 100 positions.

 Prior work gives some reason to expect this transfer to work. The
Platonic Representation Hypothesis argues that models trained on different
modalities recover increasingly similar relational structure as they scale
\cite{huh2024platonic}, while CLIP explicitly draws matching images and text
together \cite{radford2021learning}. Yet retrieval alignment does not require
identical distributions. Multimodal representation spaces have been shown to
exhibit a modality gap, with text and visual embeddings retaining
modality-specific offsets and geometry despite being mapped into a shared
space \cite{liang2022mind,zhang2024connect}. As a result, a model trained only on text embeddings may struggle to process visual embeddings, since they occupy different regions of the shared representation space.

\begin{figure}[t]
    \centering
    \includegraphics[trim=0 140 0 0,clip,width=0.94\linewidth]{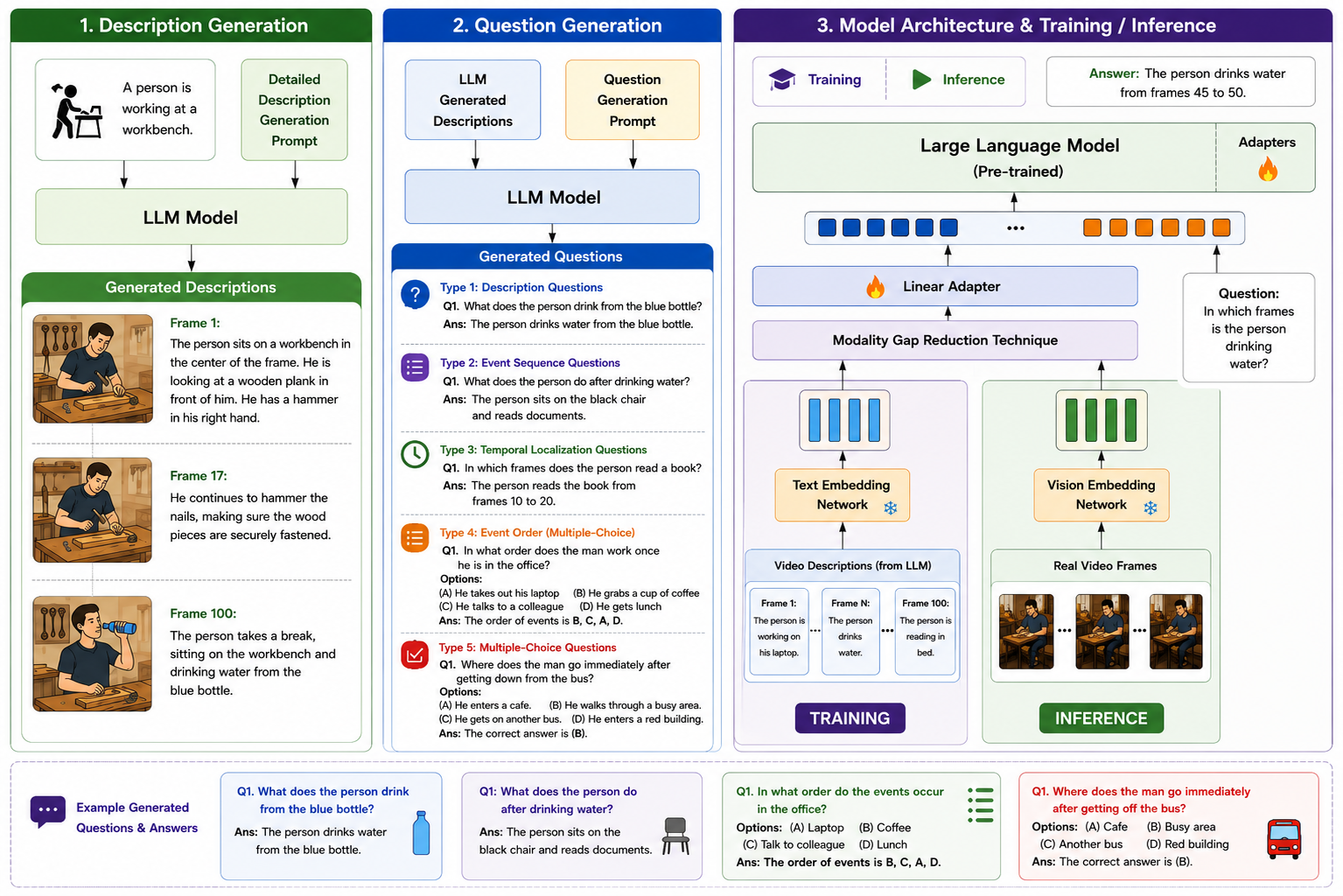}
    \caption{(a) Generation of frame-level textual sequences and questions.
    (b) Training from text embeddings and inference from video frames, with an
    optional modality-gap reduction step.}
    \label{fig:model_diagram}
\end{figure}

The text encoder comes with a separate challenge. CLIP accepts at
most 77 text tokens, with an effective context that can be considerably
shorter \cite{zhang2024longclip}. Even descriptions that fit must compress
several objects, actions, and relations into one vector. LLM2CLIP is designed
to retain richer language information \cite{huang2024llm2clip}. Because existing
video captions are typically short and at the clip level, we generate detailed and simplified views of the same descriptions to quantify the value of detail and test how existing
multimodal encoders represent each.

Our experiments use 134K synthetically generated training examples and evaluate zero-shot performance on Charades-STA \cite{gao2017tall} and QV-Highlights \cite{lei2021detecting}, without using any videos from either benchmark to train the temporal-localization model.
We first examine whether the base LLMs can localize events from
textual sequences and find that they are capable of doing so
(Table~\ref{tab:model_comparison_simple_caption}). The next challenge
is transferring this ability to video, where textual descriptions
are replaced by visual embeddings. We find that the choice of
encoder plays an important role in this transfer. While CLIP
struggles with detailed descriptions, LLM2CLIP represents them
more effectively, achieving 61.54 mtIoU compared to 53.86 on
Temporal-TextVid, and also transfers better to video
(Table~\ref{tab:caption_embedding_comparison}). Based on these
results, we adopt LLM2CLIP with detailed descriptions as our
main configuration. However, the gap between text and visual
representations remains substantial. When evaluated directly
on video, the text-trained model achieves only 4.99 mtIoU
on Charades-STA and 11.12 mtIoU on QV-Highlights. This motivates
us to explore methods for bridging the modality gap. We evaluate
four families of approaches and introduce Nearest Entity
Replacement (NER), which retrieves fine-grained text entities
for each video frame. NER improves performance to 14.45 mtIoU
on Charades-STA and 20.72 mtIoU on QV-Highlights
(Table~\ref{tab:combined_model_performance}). While these gains
show that the modality gap can be reduced, substantial errors
in temporal localization remain.
Our contributions are:
\begin{itemize}
    \item  We introduce text-only training for video temporal grounding, using
    sequences of up to 100 textual time steps. To our knowledge, this is the first
    work to train a temporal-localization model entirely on text, starting from a
    text-only LLM rather than a video-aligned one, and transfer it to video without
    visual fine-tuning or test-time adaptation.
    \item We study two challenges that arise in text-to-video transfer: preserving fine-grained descriptions in fixed-dimensional embeddings and the modality gap encountered when replacing the text embeddings used during training with visual embeddings at inference.
    \item We propose Nearest Entity Replacement (NER), a training-free mapping that
    replaces each frame embedding with the mean of its nearest text-entity
    embeddings (Eq.~\ref{eq:ner}). Across the matched encoder and caption configurations, NER improves mtIoU over raw visual features and caption-memory projection on both video benchmarks.
\end{itemize}

\section{Related Work}
\paragraph{Representation convergence and cross-modal gaps.}
Joint embedding models make cross-modal transfer possible by preserving
semantic neighborhoods across modalities. CLIP learns this structure through
image-text contrastive training \cite{radford2021learning}, and the Platonic
Representation Hypothesis places it within a broader trend toward convergent
representations across models and modalities \cite{huh2024platonic}. However,
semantic agreement does not mean that a downstream model can use raw text and
visual embeddings in the same way. Contrastively aligned modalities can
occupy distinct regions of the shared space
\cite{liang2022mind}. C3 models part of this difference with a modality-specific
mean shift \cite{zhang2024connect}, while recent work finds that the remaining
gap varies across the embedding space and cannot be corrected with a single
average shift \cite{yu2026anisoalign}.
We test this directly by training a temporal model on the text encoder's
outputs and applying it to the visual encoder's outputs without retraining.
\paragraph{Cross-modal learning from a single modality.}
Shared spaces have enabled cross-modal tasks to be learned without paired
examples for the target task. Noise-injected text training transfers an image
captioning decoder from CLIP text embeddings to image embeddings
\cite{nukrai2022text}, while DeCap first projects image embeddings through a
text memory \cite{li2023decap}. LinCIR learns composed image retrieval from
language alone \cite{gu2024language}. The reverse direction is also possible:
LAFITE trains text-to-image generation without text by constructing
text-space conditions from images \cite{zhou2022towards}. C3 applies
modality-specific centering and corruption across captioning and generation
tasks \cite{zhang2024connect}. For video, TOPA pre-aligns a base LLM with
textual videos before multiple-choice evaluation \cite{li2024topa}. T3
\cite{li2025t3} instead transfers temporal reasoning learned in text to an
already multimodal Video-LLM, and its grounding questions classify an event as
occurring at the beginning, middle, or end rather than predicting its boundaries.
TD-VAD similarly trains on text sequences and applies the model to CLIP frame
features, but for query-free anomaly detection, while concurrent captioning work
synthesizes video latents from text \cite{zhang2026tdvad,fu2026watching}. We
instead train a Video-LLM for temporal localization entirely from
text sequences and introduce video only at inference.

\paragraph{Video temporal grounding.}
Video-LLMs usually learn temporal grounding from paired video and language,
and they differ mainly in how they represent time: discrete temporal tokens
with multi-stage video training \cite{wang2025grounded}, numbered frames
\cite{wu2025numberit}, timestamp embeddings with token compression
\cite{guo2025vtgllm}, or sequential transfer from images and short clips
\cite{deng2025seq2time}. Other work improves the training data and
objectives through grounding datasets, reinforcement fine-tuning, and
comparisons of temporal output formats
\cite{chen2025datasets,li2025tempsamp,jin2026outputtime}. A second line of
work avoids task-specific annotation. Language-free grounding learns from
unlabelled videos and visual pseudo-queries \cite{kim2023languagefree}, and
zero-shot methods apply frozen visual or multimodal models directly to test
videos through temporal coherence, visual markers, structured reasoning,
self-similarity, or clip-level recognition
\cite{luo2024frozen,zheng2024trainingfree,xu2025momentgpt,jeon2026granalign,lee2025tag,fang2026markit,zhang2026dart,lee2026selfsims,li2026reze,huang2026fvaction}.
Zero-shot temporal action localization targets unseen action classes and may
still use video supervision or per-video test-time adaptation
\cite{nag2022stale,liberatori2024t3al,susladkar2024grizal,liberatori2026tegu}.
None of these approaches learns the temporal-grounding model itself from text
alone.

\section{Method}

\subsection{Problem formulation and text-only training}
For a sequence of \(T\) sampled positions, let
\(D=(d_1,\ldots,d_T)\) denote the frame-level textual descriptions, \(q\) a
query, and \(a=(a_1,\ldots,a_L)\) its answer. In temporal-localization
examples, \(a\) contains the start and end indices of the queried event. Let
\(f_T\) and \(f_V\) be the text and vision encoders of a pretrained multimodal
model, both mapping into \(\mathbb{R}^{d}\). A learned projection
\(\mathcal{P}:\mathbb{R}^{d}\rightarrow\mathbb{R}^{h}\) maps their outputs to
the token dimension of an LLM \(g\).

Our model follows TOPA's text-only pre-alignment architecture
\cite{li2024topa}. We adapt Qwen3-8B with LoRA \cite{qwen3_2025,hu2022lora},
using sequences of up to 100 positions for query-conditioned temporal
localization. The original backbone weights and multimodal encoders remain
frozen, and no Video-LLM checkpoint is used. A trainable linear projection
maps each 768-dimensional CLIP or 1,280-dimensional LLM2CLIP feature to the
4,096-dimensional language-model space, and a learned temporal embedding
\(\boldsymbol{\tau}_t\) marks each position. Training updates the projection,
temporal embeddings, and rank-16 LoRA matrices on the query, key, value, and
output projections of the attention layers, with LoRA alpha 32 and zero
dropout. Features are normalized before the learned projection.

During training, each textual position becomes
\(\mathbf{h}^{T}_t=\mathcal{P}(f_T(d_t))+\boldsymbol{\tau}_t\). The sequence
\((\mathbf{h}^{T}_1,\ldots,\mathbf{h}^{T}_T)\), followed by the tokenized
query, conditions the autoregressive answer:
\begin{equation}
\mathcal{L}
=-\sum_{\ell=1}^{L}
\log p_g\!\left(a_\ell\mid
\mathbf{h}^{T}_{1:T},q,a_{<\ell}\right).
\label{eq:training-objective}
\end{equation}
At inference, a video \(V=(F_1,\ldots,F_T)\) is sampled at the same number of
positions and supplies
\(\mathbf{h}^{V}_t=\mathcal{P}(f_V(F_t))+\boldsymbol{\tau}_t\). No parameter
is updated at inference. This lets us measure how well a model trained on the
outputs of \(f_T\) transfers to the outputs of \(f_V\).

\subsection{Synthetic training corpus}
To replace video frames with textual descriptions while preserving frame-level details and temporal information, we need a description for each sampled frame rather than a single description for the entire video. Existing video-caption datasets
generally describe clips or coarse segments, so we generate  Temporal-TextVid from
TextVid \cite{li2024topa}. As illustrated in
Fig.~\ref{fig:model_diagram}(a), Qwen3.8-27B \cite{qwen38} expands each
source description into a plausible sequence of up to 100 detailed textual
time steps. It also generates questions from five families: description,
temporal localization, event ordering, event-order multiple choice, and
general multiple choice.
Temporal-TextVid contains 134K examples. On average, each example has 29 questions
across the five families, giving 3.9M generated questions. 

\paragraph{Caption simplification.}
Existing video captions are typically short, so we generate two views of every
sequence to quantify the value of detail and to test how existing multimodal
encoders represent it. Detailed descriptions preserve more scene information,
but they place more content into a single fixed-dimensional embedding and can
exceed the text encoder's input length. For the simplified view, Qwen3.8-27B shortens
each detailed description while retaining its principal events and temporal
order. Both views therefore contain the same 134K underlying examples rather
than two independently sampled corpora.
Appendix Fig.~\ref{fig:captions_example} shows both views.

\subsection{Modality-gap reduction baselines}
\paragraph{Gaussian noise injection.}
We independently perturb each text embedding during training,
\(\widetilde{\mathbf{e}}^T_t=f_T(d_t)+\boldsymbol{\epsilon}_t\), where
\(\boldsymbol{\epsilon}_t\sim
\mathcal{N}(\mathbf{0},\sigma^2\mathbf{I})\). The perturbation exposes the
downstream model to randomly modified versions of each text embedding. The
same zero-mean Gaussian distribution is used for every caption, so the
perturbation does not depend on caption content. No noise is applied at
inference. This baseline follows prior
single-modality cross-modal training
\cite{nukrai2022text,zhang2024connect,zhou2022towards}.

\paragraph{Collapse and corrupt.}
Let \(\mathcal{D}_T\) be the training descriptions and
\(\mathcal{D}_V\) a reference set of visual inputs. C3 estimates one mean for
each modality:
\begin{equation}
\boldsymbol{\mu}_T
=\frac{1}{|\mathcal{D}_T|}
\sum_{d\in\mathcal{D}_T}f_T(d),
\qquad
\boldsymbol{\mu}_V
=\frac{1}{|\mathcal{D}_V|}
\sum_{F\in\mathcal{D}_V}f_V(F).
\label{eq:modality-means}
\end{equation}
C3 centers and corrupts text during training,
\(\widetilde{\mathbf{e}}^T_t=f_T(d_t)-\boldsymbol{\mu}_T+
\boldsymbol{\epsilon}_t\), and centers visual inputs at inference,
\(\widetilde{\mathbf{e}}^V_t=f_V(F_t)-\boldsymbol{\mu}_V\)
\cite{zhang2024connect}. With a fixed visual reference, the same pair of means is
used for every sample, so C3 treats the modality difference as a single dataset-level offset. We
compare four estimates of
\(\boldsymbol{\mu}_V\): the current video, the complete inference split, a
similarly sized video dataset, and a large LAION reference sample. Appendix
Table~\ref{tab:combined_performance} reports the full sensitivity analysis.

\paragraph{Mixed-modality training.} 
We encode approximately six million LAION-400M images
\cite{schuhmann2021laion} with the vision encoder to form a visual memory.
After 35 epochs, the curriculum selects 10\% of training examples for
replacement. Within a selected example, the variants replace a subset of
positions with a nearest visual feature or an average of 10 visual neighbors,
or replace every position. These experiments test whether exposure to retrieved visual vectors prepares the
LLM for video inference.

\paragraph{Caption-memory projection.}
The DeCap/TOPA baseline \cite{li2023decap,li2024topa} stores caption
embeddings in a support memory
\(\mathbf{M}_C=\{\mathbf{c}_i\}_{i=1}^{K_C}\). A visual embedding
\(\mathbf{v}\) is mapped back to the text space by a
temperature-controlled weighted average:
\begin{equation}
\widehat{\mathbf{v}}_C
=\sum_{i=1}^{K_C}
\frac{\exp(\mathbf{c}_i^\top\mathbf{v}/\lambda)}
{\sum_{j=1}^{K_C}\exp(\mathbf{c}_j^\top\mathbf{v}/\lambda)}
\mathbf{c}_i.
\label{eq:projection-decoding}
\end{equation}
We refer to this as caption-memory projection in the experiments.

\subsection{Nearest Entity Replacement}
NER can be viewed as soft semantic quantization. A frame embedding selects a
set of nearby text entities, and their mean replaces the visual vector. The
mapping is soft because each frame is represented by an average over many
entities rather than one discrete entry. Unlike a single global offset, this
replacement depends on the frame content and gives the temporal model the same
type of text representation used during training.

\begin{figure}[t]
\centering
\includegraphics[width=0.72\linewidth]{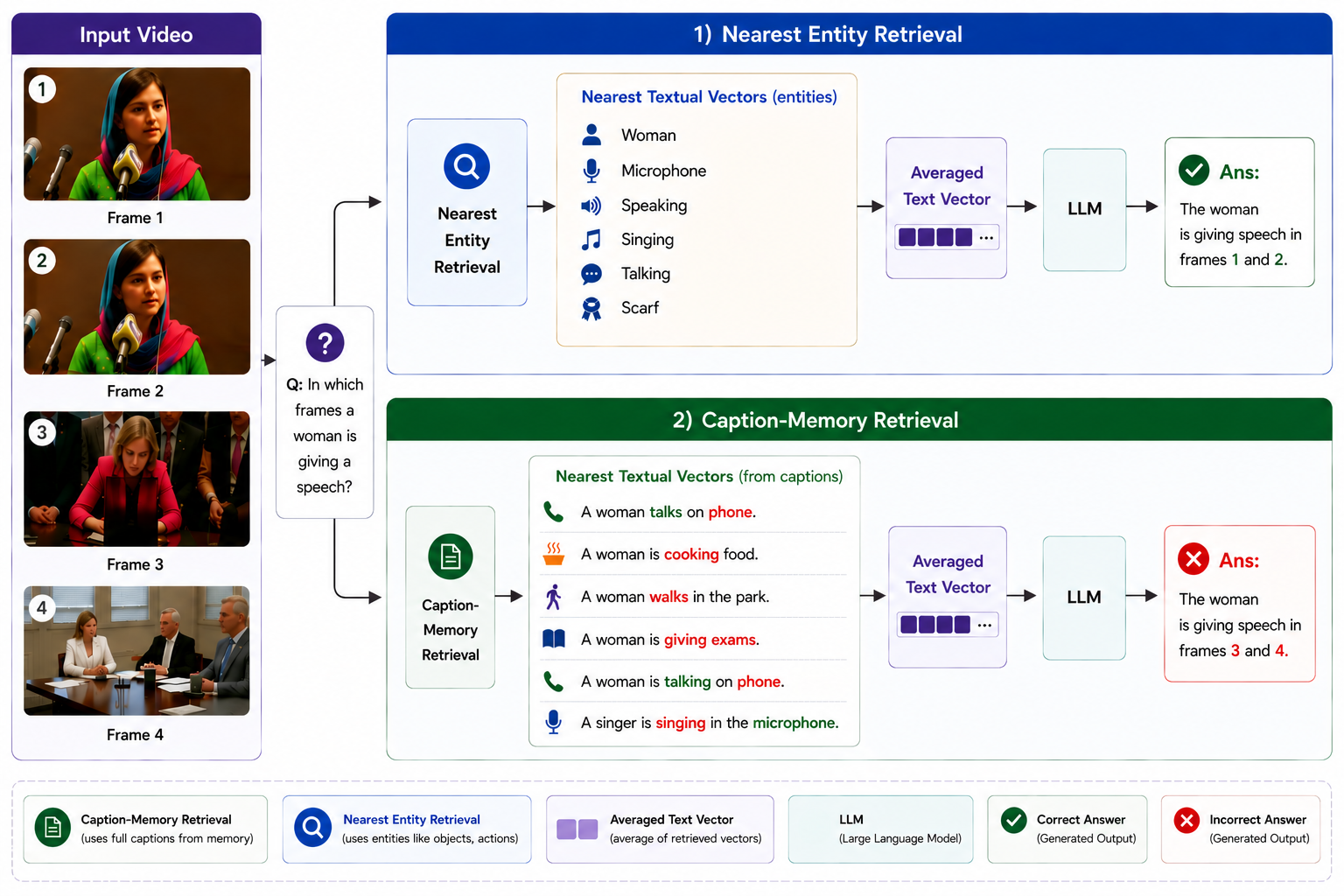}
\caption{NER avoids concept bleed from unrelated caption content (red).}
\label{fig:entityl}
\end{figure}

\paragraph{Entity extraction and memory.}
A caption-memory entry can bind several concepts into one vector. For example,
``a woman is sitting on a chair and working at a table'' combines the woman,
chair, sitting, working, and table. A frame containing only one of these
concepts may retrieve the complete caption and introduce the others
(Fig.~\ref{fig:entityl}). We call this \emph{concept bleed}: the matching part
of a caption brings unrelated caption content into the frame representation.
NER instead extracts nouns, verbs, and action phrases from the training
descriptions. It encodes each phrase separately with the text encoder and
stores the resulting 200K entries in an entity memory
\(\mathbf{M}_E=\{\mathbf{m}_i\}_{i=1}^{K_E}\).

\paragraph{Frame-to-entity retrieval.}
For each frame \(F_t\), NER obtains
\(\mathbf{v}_t=f_V(F_t)\), selects the \(k\) entity vectors with the highest
cosine similarity, and records their indices in \(\mathcal{N}_k(t)\).

\paragraph{Aggregation and replacement.}
NER replaces the visual vector with the mean of the retrieved text entities:
\begin{equation}
\mathcal{N}_k(t)=\operatorname{TopK}_{i}\,
\frac{\mathbf{v}_t^\top\mathbf{m}_i}
{\|\mathbf{v}_t\|\,\|\mathbf{m}_i\|},
\qquad
\widehat{\mathbf{v}}^{\,E}_t
=\operatorname{Mean}\{\mathbf{m}_i : i\in\mathcal{N}_k(t)\}.
\label{eq:ner}
\end{equation}
The averaging makes the replacement less dependent on any single cross-modal
match. The LLM then receives
\(\{\mathcal{P}(\widehat{\mathbf{v}}^{\,E}_t)+
\boldsymbol{\tau}_t\}_{t=1}^{T}\). The visual embedding is used only to
retrieve a text-space neighborhood, and no video-side parameter is updated.

\section{Experimental Setup}
\paragraph{Models and protocol.}
All trained temporal models use Qwen3-8B with LoRA and Temporal-TextVid as
their training data. CLIP and LLM2CLIP provide the paired text and vision
encoders. The main raw-input, caption-projection, and NER configuration uses
LLM2CLIP with detailed descriptions. Gaussian, C3, and mixed-modality training
use CLIP ViT-L/14 with simplified descriptions, and we also report NER with this
CLIP configuration.
We evaluate text-side behavior on
Temporal-TextVid and video transfer on Charades-STA and the QV-Highlights
validation split. No videos from either benchmark are used for training. NER's
neighbor count \(k\) is selected separately for each encoder and caption view on
labelled development videos from each benchmark's training split
(Appendix~\ref{sec:impl}). The main configuration uses \(k=1{,}000\) on
Charades-STA and \(k=20\) on QV-Highlights, and NER with CLIP uses
\(k=4{,}000\) and \(k=1{,}000\).
Training uses 100 epochs, an effective batch size of 1,536, AdamW (peak
learning rate \(6\times10^{-4}\), one warmup epoch, cosine decay), and seed 0.
Checkpoints are selected on a separate text monitor.

\paragraph{Metrics.}
Temporal intersection over union (tIoU) divides the intersection of the
predicted and reference intervals by their union. We report its dataset mean
(mtIoU) and the percentage of examples reaching tIoU thresholds of \(0.1\),
\(0.3\), and \(0.5\). All reported values are on a 0--100 scale.
We use the archived preprocessing and all-row scoring (Appendix~\ref{sec:impl}): Charades-STA has
2,572 query--interval examples, and QV-Highlights averages over 2,701
query--reference-window rows (1,486 queries) rather than the best reference per
query. Published rows retain their original protocols.

\section{Results}
\subsection{Temporal localization from text}
Before replacing text with video, we test whether frozen Qwen3-8B, prompted
without LoRA, and Qwen3.8-27B can localize events directly from Temporal-TextVid descriptions.
Each prompt either labels descriptions with frame
indices or presents them in order without explicit indices. This isolates
temporal reasoning from losses introduced by the embedding encoder and the
text-to-video switch. Since Qwen3.8-27B generated the corpus, its score measures
consistency with the synthetic descriptions rather than independent
generalization.

Frame indices are important for both models
(Table~\ref{tab:model_comparison_simple_caption}). Removing them lowers
Qwen3-8B's mtIoU from 45.47 to 13.82 and Qwen3.8-27B's from 55.69 to 28.72. This suggests that neither model can reliably identify frame positions from sequence order alone.

\subsection{Effect of caption detail}
We next compress each description into one embedding and compare CLIP with
LLM2CLIP (Table~\ref{tab:caption_embedding_comparison}). 

\begin{table}[t]
\centering
\begin{minipage}[t]{0.52\linewidth}
\vspace{0pt}
\caption{Direct text localization with and without frame IDs on Temporal-TextVid.}
\label{tab:model_comparison_simple_caption}
\centering
\scriptsize
\renewcommand{\arraystretch}{0.9}
\setlength{\tabcolsep}{2.2pt}
\begin{tabular}{@{}lcrrrr@{}}
\toprule
Model & Frame IDs & mtIoU & R@0.1 & R@0.3 & R@0.5 \\
\midrule
Qwen3-8B & Yes & 45.47 & 64.71 & 57.91 & 41.48 \\
Qwen3-8B & No  & 13.82 & 25.81 & 20.46 & 11.47 \\
Qwen3.8-27B & Yes & 55.69 & 68.44 & 64.88 & 53.84 \\
Qwen3.8-27B & No  & 28.72 & 42.02 & 36.89 & 27.02 \\
\bottomrule
\end{tabular}
\end{minipage}
\hfill
\begin{minipage}[t]{0.46\linewidth}
\vspace{0pt}
\caption{Caption-view and embedding-model ablation (mtIoU). TTV denotes
Temporal-TextVid.}
\label{tab:caption_embedding_comparison}
\centering
\scriptsize
\renewcommand{\arraystretch}{0.9}
\setlength{\tabcolsep}{1.7pt}
\begin{tabular}{@{}llrrr@{}}
\toprule
View & Embedding & TTV & Char. & QV-H. \\
\midrule
Detailed   & CLIP     & 53.86 & 2.64 & 3.29 \\
Detailed   & LLM2CLIP & 61.54 & 4.99 & 11.12 \\
Simplified & CLIP     & 56.64 & 7.02 & 9.25 \\
Simplified & LLM2CLIP & 59.34 & 5.61 & 10.83 \\
\bottomrule
\end{tabular}
\end{minipage}
\end{table}

For CLIP, simplifying the descriptions raises Temporal-TextVid mtIoU from 53.86
to 56.64, although both views contain the same examples. The detailed view places more objects, actions, and relations into each fixed-dimensional
representation. LLM2CLIP instead scores higher with detailed descriptions
(61.54 versus 59.34), consistent with its LLM-based text encoder accepting
longer inputs than CLIP's 77 tokens \cite{huang2024llm2clip}. 

On video, CLIP fails with detailed descriptions: its raw transfer falls from
7.02 to 2.64 on Charades-STA and from 9.25 to 3.29 on QV-Highlights. LLM2CLIP
is much less sensitive to the caption view, moving from 5.61 and 10.83 to 4.99
and 11.12. We therefore keep detailed descriptions with LLM2CLIP, which gives
the best text score and avoids CLIP's failure on video. Even so, these scores
fall far below its 61.54 text score, highlighting the substantial domain gap.

\begin{figure}[t]
\centering
\begin{minipage}[t]{0.41\linewidth}
\vspace{0pt}
\captionof{table}{C3 sensitivity to the visual-mean reference with CLIP
(mtIoU).}
\label{tab:c3_mtiou}
\centering
\scriptsize
\setlength{\tabcolsep}{3pt}
\begin{tabular}{@{}lrr@{}}
\toprule
Reference & Char. & QV-H. \\
\midrule
Current video    & \textbf{5.78} & \textbf{11.55} \\
Evaluation split & 5.09 & 10.76 \\
Other benchmark  & 4.74 & 7.31 \\
LAION (5.99M)    & 5.10 & 10.16 \\
\bottomrule
\end{tabular}
\end{minipage}
\hfill
\begin{minipage}[t]{0.57\linewidth}
\vspace{0pt}
\centering
\includegraphics[width=0.88\linewidth]{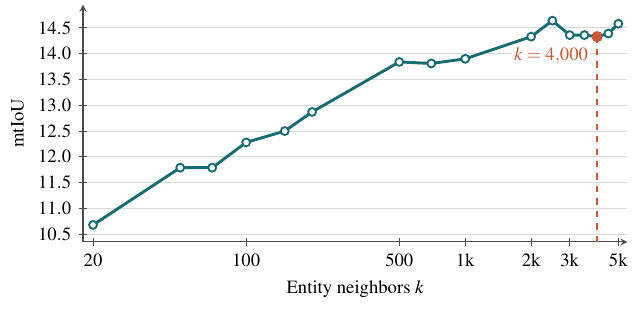}
\caption{NER (CLIP) on Charades-STA across neighbor counts, with the selected
\(k=4{,}000\) marked.}
\label{fig:ner_neighbors}
\end{minipage}
\end{figure}

\subsection{Improving text-to-video transfer}
Table~\ref{tab:combined_model_performance} compares NER with four families of
methods for closing the text-to-video gap. The training-side baselines use
CLIP. Gaussian noise injection and C3 alter the text embeddings independently
of frame content and reach only 5.46 and 5.10 mtIoU on Charades-STA, and C3 is
also sensitive to the visual reference used for its mean
(Table~\ref{tab:c3_mtiou}). Mixing retrieved image features into training helps more and gives the
strongest training baselines, with 9.45 mtIoU on Charades-STA for partial and
12.74 on QV-Highlights for full replacement (Appendix Table~\ref{tab:mixed_modality_full}).
At inference, caption-memory projection raises the detailed LLM2CLIP model from
4.99 to 7.95 mtIoU on Charades-STA and from 11.12 to 19.01 on QV-Highlights.
\begin{table}[!t]
\caption{Transfer results and published context under original protocols. CLIP
rows use simplified descriptions and LLM2CLIP rows detailed descriptions. \(\dagger\):
values reproduced from VTimeLLM \cite{huang2024vtimellm}.}
\label{tab:combined_model_performance}
\centering
\scriptsize
\renewcommand{\arraystretch}{0.82}
\setlength{\tabcolsep}{2.2pt}
\begin{tabular}{@{}llllrrrr@{}}
\toprule
Dataset & Training & Method & Encoder & mtIoU & R@0.1 & R@0.3 & R@0.5 \\
\midrule
Charades & Video trained & VideoChat-7B\(\dagger\) \cite{li2023videochat} & -- & 6.50 & -- & 9.00 & 3.30 \\
 & & VideoChatGPT-7B\(\dagger\) \cite{maaz2024videochatgpt} & -- & 13.70 & -- & 20.00 & 7.70 \\
 & & VTimeLLM-7B \cite{huang2024vtimellm} & -- & 31.20 & -- & 51.00 & 27.50 \\
 & & Grounded-VideoLLM (Phi3.5, 4B) \cite{wang2025grounded} & -- & 36.80 & -- & 54.20 & 36.40 \\
\midrule
Charades & Text + image memory & Mixed, full & CLIP & 8.78 & 22.78 & 9.33 & 4.39 \\
 & Text only & Gaussian noise injection \cite{nukrai2022text} & CLIP & 5.46 & 17.19 & 3.69 & 1.01 \\
 & & C3 \cite{zhang2024connect} & CLIP & 5.10 & 15.36 & 3.46 & 1.05 \\
 & & Raw visual features & LLM2CLIP & 4.99 & 12.99 & 5.75 & 2.29 \\
 & & Caption projection \cite{li2023decap,li2024topa} & LLM2CLIP & 7.95 & 18.93 & 9.18 & 3.97 \\
 & & NER (ours) & CLIP & 14.33 & 36.39 & \textbf{20.53} & \textbf{7.08} \\
 & & NER (ours) & LLM2CLIP & \textbf{14.45} & \textbf{40.01} & 20.49 & 6.07 \\
\midrule
QV-Highlights & Text + image memory & Mixed, full & CLIP & 12.74 & 29.62 & 15.81 & 9.29 \\
 & Text only & Gaussian noise injection \cite{nukrai2022text} & CLIP & 11.14 & 26.95 & 14.18 & 7.74 \\
 & & C3 \cite{zhang2024connect} & CLIP & 10.16 & 25.51 & 12.88 & 6.85 \\
 & & Raw visual features & LLM2CLIP & 11.12 & 24.51 & 14.07 & 8.03 \\
 & & Caption projection \cite{li2023decap,li2024topa} & LLM2CLIP & 19.01 & 39.76 & 24.21 & 15.59 \\
 & & NER (ours) & CLIP & \textbf{21.85} & \textbf{45.65} & \textbf{26.69} & \textbf{17.18} \\
 & & NER (ours) & LLM2CLIP & 20.72 & 44.17 & 26.10 & 16.14 \\
\bottomrule
\end{tabular}
\end{table}

NER outperforms every baseline on both benchmarks. With detailed LLM2CLIP, it
reaches 14.45 mtIoU on Charades-STA and 20.72 on QV-Highlights, 9.46 and 9.60
points above raw visual features and 6.50 and 1.71 above caption-memory
projection, whose R@0.1 and R@0.3 it more than doubles on Charades-STA. With
CLIP, NER reaches 14.33 and 21.85, 4.88 and 9.11 points above the strongest
CLIP training baseline.

 We attribute the gain over caption projection to the retrieval unit. A caption
binds several concepts into one vector, so a frame that matches part of a
caption also inherits the rest (Fig.~\ref{fig:entityl}). Averaging many
single-concept entities instead gives a vector that follows the frame content
while staying in the text distribution seen during training. Across encoder and
caption configurations, simplified LLM2CLIP + NER scores highest, at 15.57 and
22.29 (Appendix Table~\ref{tab:matched_retrieval}). However, models trained on paired video-text data still achieve up to 36.80 mtIoU under their original protocols, highlighting the performance gap between single-modality and paired-data approaches.
\subsection{Sensitivity to the entity-neighbor count}
 The neighbor count \(k\) sets how many entities are averaged for each frame,
and larger neighborhoods average out individual wrong matches. For CLIP + NER
on Charades-STA
(Fig.~\ref{fig:ner_neighbors}), mtIoU rises from 10.68 at \(k=20\) to 14.33 at
the selected \(k=4{,}000\) and varies by less than 0.4 points beyond \(k=2{,}000\), so performance is
stable over a wide range of \(k\). Development and confirmation results for
every configuration are in Appendix~\ref{sec:impl}.

\section{Discussion and Limitations}
Contrastive training aligns image and text embeddings well enough for retrieval, but our experiments show that this does not make them interchangeable. A model that localizes events well from text loses most of its accuracy when frame embeddings replace the descriptions, even with the stronger LLM2CLIP encoder.

We believe our simple method, NER, works because the temporal model never directly processes visual embeddings. Instead, frame embeddings are used to retrieve nearby text entities, allowing the model to work with the same type of input it was trained on. This lets NER benefit from the retrieval capabilities of existing encoders without requiring their image and text embedding spaces to be perfectly aligned. In contrast, Gaussian noise and C3 apply the same adjustment to every input regardless of frame content, while C3 also depends on its visual reference. This is consistent with evidence that the modality gap varies across the embedding space \cite{yu2026anisoalign}. Caption-memory projection accounts for individual frames but can introduce information from captions that is not actually present in the frame.

NER is still better at identifying the general region of an event than its precise boundaries, leaving several directions for improvement. Retrieved entities could be weighted by similarity rather than averaged equally, or the number of neighbors could be adjusted for each frame. The mapping could also be learned from unpaired frames and text, while the synthetic corpus could better reflect the timing and duration of events in real videos. Better-aligned encoders may eventually allow direct transfer without retrieval. Finally, evaluation on more diverse benchmarks is needed to understand how well these findings generalize.
\clearpage
{\small
\bibliographystyle{plainnat}
\bibliography{references}
}

\clearpage
\appendix
\section{Additional Results}

\subsection{Published QV-Highlights Video-LLM results}
Video-LLMs commonly evaluate QV-Highlights as highlight detection, using mAP
and HIT@1 rather than the interval metrics reported for NER. Table
\ref{tab:qv_highlight_context} therefore presents these results separately.

\begin{table}[!ht]
\caption{Published QV-Highlights highlight-detection results. The first three
values are reproduced from VTG-LLM \cite{guo2025vtgllm} and NumPro-FT from Number It
\cite{wu2025numberit}. NumPro-FT uses video fine-tuning, and the other rows are zero-shot. These saliency
metrics are not directly comparable to NER.}
\label{tab:qv_highlight_context}
\centering
\small
\begin{tabular}{@{}lrr@{}}
\toprule
Method & mAP & HIT@1 \\
\midrule
VideoChat-7B \cite{li2023videochat} & 13.1 & 18.1 \\
TimeChat-7B \cite{ren2024timechat} & 14.5 & 23.9 \\
VTG-LLM-7B \cite{guo2025vtgllm} & 16.5 & 33.5 \\
NumPro-FT \cite{wu2025numberit} & 25.0 & 37.2 \\
\bottomrule
\end{tabular}
\end{table}
\FloatBarrier

\subsection{Mixed-modality variants}
\begin{table}[!ht]
\caption{Full mixed-modality training results with CLIP.}
\label{tab:mixed_modality_full}
\centering
\small
\begin{tabular}{@{}llrrrr@{}}
\toprule
Dataset & Variant & mtIoU & R@0.1 & R@0.3 & R@0.5 \\
\midrule
Charades & Partial & 9.45 & 21.77 & 11.86 & 5.56 \\
 & Partial + neighbors & 7.07 & 17.19 & 8.20 & 3.65 \\
 & Full & 8.78 & 22.78 & 9.33 & 4.39 \\
\midrule
QV-Highlights & Partial & 12.41 & 26.47 & 15.66 & 9.55 \\
 & Partial + neighbors & 9.35 & 22.25 & 11.48 & 6.26 \\
 & Full & 12.74 & 29.62 & 15.81 & 9.29 \\
\bottomrule
\end{tabular}
\end{table}
\FloatBarrier

\subsection{Sensitivity to the C3 visual mean}
 C3 requires a visual mean at inference. Table~\ref{tab:combined_performance} compares a current-video estimate, a transductive estimate from the full evaluation split, a different video dataset, and a 5.99M-image LAION sample. The current-video estimate gives the highest mtIoU on both datasets. LAION is the fixed C3 configuration in Table~\ref{tab:combined_model_performance}. Neither reference is uniformly best across recall thresholds: on Charades-STA, the evaluation-split estimate has higher R@0.3 and R@0.5 than both. A different video dataset of similar size performs worst on both benchmarks, so the source of the reference matters.

\begin{table}[t]
\caption{C3 sensitivity to the visual-mean reference with CLIP. The evaluation-split mean uses the full target split and is therefore transductive. The other video set is the other benchmark.}
\label{tab:combined_performance}
\centering
\small
\setlength{\tabcolsep}{4pt}
\begin{tabular}{@{}llrrrr@{}}
\toprule
Dataset & Visual-mean reference & mtIoU & R@0.1 & R@0.3 & R@0.5 \\
\midrule
Charades & Current video & \textbf{5.78} & \textbf{18.74} & 3.34 & 1.17 \\
 & Eval split (133.4K) & 5.09 & 14.74 & \textbf{3.50} & \textbf{1.28} \\
 & Other video set (145.5K) & 4.74 & 14.15 & 2.95 & 0.89 \\
 & LAION (5.99M) & 5.10 & 15.36 & 3.46 & 1.05 \\
\midrule
QV-Highlights & Current video & \textbf{11.55} & \textbf{28.40} & \textbf{15.29} & \textbf{7.77} \\
 & Eval split (145.5K) & 10.76 & 26.14 & 14.07 & 7.18 \\
 & Other video set (133.4K) & 7.31 & 19.62 & 9.07 & 4.07 \\
 & LAION (5.99M) & 10.16 & 25.51 & 12.88 & 6.85 \\
\bottomrule
\end{tabular}
\end{table}

Table~\ref{tab:caption_embedding_full} reports all recall thresholds for the
caption-view and encoder comparison in Table~\ref{tab:caption_embedding_comparison}.

\begin{table}[t]
\caption{Full caption-view and embedding-model results. TTV denotes Temporal-TextVid.}
\label{tab:caption_embedding_full}
\centering
\small
\setlength{\tabcolsep}{3pt}
\begin{tabular}{@{}lllrrrrr@{}}
\toprule
View & Embedding & Dataset & mtIoU & R@0.1 & R@0.3 & R@0.5 & R@0.7 \\
\midrule
Detailed & CLIP & TTV & 53.86 & 69.78 & 65.72 & 51.85 & 43.93 \\
 & & Charades & 2.64 & 6.53 & 3.27 & 0.93 & 0.62 \\
 & & QV-Highlights & 3.29 & 8.66 & 3.30 & 1.59 & 0.78 \\
\addlinespace
Detailed & LLM2CLIP & TTV & 61.54 & 76.18 & 72.68 & 60.18 & 52.31 \\
 & & Charades & 4.99 & 12.99 & 5.75 & 2.29 & 1.01 \\
 & & QV-Highlights & 11.12 & 24.51 & 14.07 & 8.03 & 4.96 \\
\addlinespace
Simplified & CLIP & TTV & 56.64 & 73.18 & 67.95 & 54.39 & 46.27 \\
 & & Charades & 7.02 & 16.25 & 8.98 & 4.00 & 2.33 \\
 & & QV-Highlights & 9.25 & 20.84 & 11.14 & 6.92 & 4.63 \\
\addlinespace
Simplified & LLM2CLIP & TTV & 59.34 & 74.66 & 70.59 & 57.34 & 50.13 \\
 & & Charades & 5.61 & 12.79 & 6.42 & 3.30 & 1.98 \\
 & & QV-Highlights & 10.83 & 24.44 & 14.55 & 7.92 & 4.44 \\
\bottomrule
\end{tabular}
\end{table}

\FloatBarrier
\subsection{Matched encoder and retrieval comparison}
 Table~\ref{tab:matched_retrieval} applies raw features, caption projection
(CP), and NER to each text-trained checkpoint. NER, with \(k\) selected per configuration on development videos, gives the
highest mtIoU in all eight rows.

\begin{table}[!ht]
\caption{Raw features, caption projection (CP), and NER for each encoder and
caption view (mtIoU).}
\label{tab:matched_retrieval}
\centering
\small
\begin{tabular}{@{}lllrrr@{}}
\toprule
Dataset & Encoder & View & Raw & CP & NER \\
\midrule
Charades & CLIP & Simplified & 7.02 & 5.31 & \textbf{14.33} \\
 & CLIP & Detailed & 2.64 & 3.42 & \textbf{7.67} \\
 & LLM2CLIP & Simplified & 5.61 & 9.37 & \textbf{15.57} \\
 & LLM2CLIP & Detailed & 4.99 & 7.95 & \textbf{14.45} \\
\midrule
QV-Highlights & CLIP & Simplified & 9.25 & 8.85 & \textbf{21.85} \\
 & CLIP & Detailed & 3.29 & 6.23 & \textbf{15.28} \\
 & LLM2CLIP & Simplified & 10.83 & 19.89 & \textbf{22.29} \\
 & LLM2CLIP & Detailed & 11.12 & 19.01 & \textbf{20.72} \\
\bottomrule
\end{tabular}
\end{table}

\FloatBarrier
\subsection{Implementation details}
\label{sec:impl}
\paragraph{Model and training.}  The temporal backbone is Qwen3-8B with frozen weights. LoRA (rank 16, alpha 32, dropout 0) is applied to the query,
key, value, and output projections. The linear feature projection (768-dimensional
CLIP or 1,280-dimensional LLM2CLIP features to 4,096 dimensions) and the temporal embeddings are trained, and features are normalized before the projection.
We train for 100 epochs with batch size 1,536 using AdamW (learning rate
\(6\times10^{-4}\), weight decay 0.1) with one warmup epoch and cosine decay, and select the checkpoint on a held-out text set.

\paragraph{Retrieval.} The entity memory has 200K entries and the caption
memory 500K entries. Caption projection uses \(\lambda=0.01\) and keeps its original settings.

\paragraph{Neighbor selection.} Model weights stay fixed, and only NER's \(k\) is
selected with video. For each benchmark, we use 200 source videos from the
training split for development and a disjoint 200 for confirmation. Neither overlaps the reported cohorts. For each encoder and caption view, we choose the
\(k\) with the highest development mtIoU from 16 values between 20 and 5,000.
Table~\ref{tab:ner_selection} compares the selected values with the earlier
\(k=3{,}000\) (Charades-STA) and \(k=2{,}000\) (QV-Highlights). Intervals are 95\% paired bootstrap intervals over source videos.

\begin{table}[!ht]
\caption{NER neighbor counts selected on development videos (mtIoU). Reported:
full evaluation sets. Confirmation: 200 separate videos per benchmark, with a 95\%
bootstrap interval for the change.}
\label{tab:ner_selection}
\centering
\small
\setlength{\tabcolsep}{3pt}
\begin{tabular}{@{}lllcccc@{}}
\toprule
Dataset & Encoder & View & \(k\) & Reported & Confirmation & Conf.\ 95\% CI \\
\midrule
Charades & CLIP & Simplified & 3{,}000\(\rightarrow\)4{,}000 & 14.36\(\rightarrow\)14.33 & 12.67\(\rightarrow\)13.42 & [\(-\)0.24, 1.73] \\
 & CLIP & Detailed & 3{,}000\(\rightarrow\)5{,}000 & 7.64\(\rightarrow\)7.67 & 6.63\(\rightarrow\)7.86 & [0.54, 2.02] \\
 & LLM2CLIP & Simplified & 3{,}000\(\rightarrow\)500 & 15.48\(\rightarrow\)15.57 & 16.46\(\rightarrow\)16.39 & [\(-\)2.25, 2.06] \\
 & LLM2CLIP & Detailed & 3{,}000\(\rightarrow\)1{,}000 & 13.91\(\rightarrow\)14.45 & 14.07\(\rightarrow\)15.37 & [\(-\)0.31, 2.97] \\
\midrule
QV-Highlights & CLIP & Simplified & 2{,}000\(\rightarrow\)1{,}000 & 21.20\(\rightarrow\)21.85 & 20.93\(\rightarrow\)22.02 & [\(-\)0.35, 2.56] \\
 & CLIP & Detailed & 2{,}000\(\rightarrow\)5{,}000 & 15.52\(\rightarrow\)15.28 & 16.00\(\rightarrow\)14.55 & [\(-\)3.12, 0.18] \\
 & LLM2CLIP & Simplified & 2{,}000\(\rightarrow\)100 & 18.63\(\rightarrow\)22.29 & 17.53\(\rightarrow\)21.80 & [1.99, 6.82] \\
 & LLM2CLIP & Detailed & 2{,}000\(\rightarrow\)20 & 16.00\(\rightarrow\)20.72 & 16.71\(\rightarrow\)22.70 & [3.39, 8.78] \\
\bottomrule
\end{tabular}
\end{table}

\paragraph{Evaluation.} The Temporal-TextVid evaluation contains about 2K
examples, Charades-STA 2,572 query--interval examples, and QV-Highlights 2,701
query--window rows from 1,486 queries. Scores average over all rows, and the rare unparseable predictions count as zero.

\paragraph{Compute.} All experiments ran on AWS p5.48xlarge instances with eight
NVIDIA H100 80GB GPUs each. Generating Temporal-TextVid with Qwen3.8-27B (FP8,
vLLM) took about five hours on six such instances. Each temporal model trained
on four GPUs for 20 to 26 hours, about 830 GPU-hours for the nine training runs,
and the 216 evaluations, including the neighbor-selection sweeps, took about 39
GPU-hours on single GPUs.

\FloatBarrier
\subsection{Broader impact}
Text-only training can reduce paired-video data collection. Predicted
intervals can still be incorrect, so downstream systems require task-specific
reliability evaluation.

\begin{figure}[t]
\centering
\includegraphics[width=\linewidth]{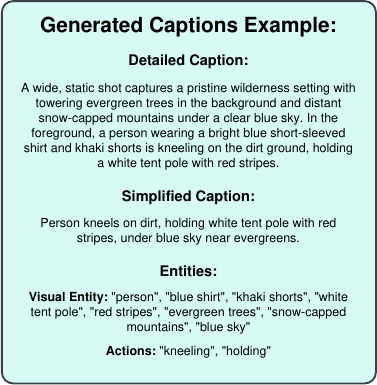}
\caption{Detailed and simplified caption views for one Temporal-TextVid example.}
\label{fig:captions_example}
\end{figure}

\end{document}